# Reconstruct Anomaly to Normal: Adversarial Learned and Latent Vector-constrained Autoencoder for Time-series Anomaly Detection

Chunkai Zhang[1], Wei Zuo[2], Xuan Wang[3]

**Abstract** Anomaly detection in time series has been widely researched and has important practical applications. In recent years, anomaly detection algorithms are mostly based on deep-learning generative models and use the reconstruction error to detect anomalies. They try to capture the distribution of normal data by reconstructing normal data in the training phase, then calculate the reconstruction error of test data to do anomaly detection. However, most of them only use the normal data in the training phase and can not ensure the reconstruction process of anomaly data. So, anomaly data can also be well reconstructed sometimes and gets low reconstruction error, which leads to the omission of anomalies. What's more, the neighbor information of data points in time series data has not been fully utilized in these algorithms. In this paper, we propose **R**AN based on the idea of **R**econstruct **A**nomalies to **N**ormal and apply it for unsupervised time series anomaly detection. To minimize the reconstruction error of normal data and maximize this of anomaly data, we do not just ensure normal data to reconstruct well, but also try to make the reconstruction of anomaly data consistent with the distribution of normal data, then anomalies will get higher reconstruction errors. We implement this idea by introducing the "imitated anomaly data" and combining a specially designed latent vector-constrained Autoencoder with the discriminator to construct an adversary network. Extensive experiments on time-series datasets from different scenes such as ECG diagnosis also show that RAN can detect meaningful anomalies, and it outperforms other algorithms in terms of AUC-ROC.

✉ Chunkai Zhang
   ckzhang@hit.edu.cn
   Wei Zuo
   zuoweijx1997@gmail.com
   Xuan Wang
   wangxuan@cs.hitsz.edu.cn

   Department of Computer Science and Technology, Harbin Institute of Technology (Shenzhen), Shenzhen, China

## 1 Introduction

Anomaly detection in time series has been studied in many fields and has important practical meanings. For example, Anomaly subsequences in electrocardiograms can indicate the health status of the heart [1, 2], anomaly financial transaction data can be credit card fraud [3], and network anomaly detection can protect the system from attacks [4]. Since anomalies always have different types and labeling real-life data are usually difficult, anomaly detection algorithms are generally unsupervised.

There are two main unsupervised anomaly detection methods according to whether using neural networks. The first one is the classical methods which generally use different ways to represent time series data, then calculates the similarity of subsequences as anomaly scores. However, most of these algorithms may lose important information in dimension reduction and suffer from the "curse of dimension" in similarity calculation [5, 6]. As the volume and dimension of time-series data grow, the classical methods become powerless to handle these complex data. More and more deep-learning methods [7] are proposed for anomaly detection, and they use neural networks to extract features automatically and calculate anomaly scores based on these features.

In recent years, deep-learning anomaly detection algorithms are mostly based on generative models [8, 9, 10, 11]. The main idea of these algorithms is that anomaly data can not be well reconstructed, while normal data can be well reconstructed by the generative model. In the training phase, these algorithms attempt to extract features of normal data by reconstructing normal data as possible. Then, in the testing phase, they calculate the reconstruction error of test data and detect samples with high error as anomalies. However, this idea is not rigorous and can lead to the omission of anomalies. Abnormal samples can also be reconstructed well sometimes [12] since abnormal samples have not been seen by the generative model and the corresponding reconstruction process is unrestrained and unknown. What's more, most of these deep-learning algorithms are fit for image anomaly detection and may not perform well in time-series, since they have not fully utilized the temporal correlation of time series data.

To fill the above drawbacks, we proposed a new deep-learning algorithm **RAN** based on the idea of **R**econstruct **A**nomalies to **N**ormal for unsupervised time series anomaly detection. First, we proposed a new training strategy to make the model see not only normal data but also anomaly data. Considering that the anomaly in an abnormal subsequence is usually a part rather than the whole, we corrupt the normal subsequences to imitate anomaly subsequences and use both of them as input for model training. Inspired by the success of the generative model and adversarial learning, we migrate and improve the architecture of [13], and then proposed **RAN** as the backbone of our solution. To utilize the temporal correlation in the subsequence, we use the fully convolutional network with different kernel sizes to construct Autoencoder, then we can extract rich neighbor information. Opposed to [13], we hope the reconstruction of anomaly data is similar to this of normal data since we are based on reconstruction errors to detect anomalies. So, we add constrain on the latent vector of corrupted subsequences and that of normal subsequences to be the same as possible, which force the encoder to learn robust features of normal subsequences and generate normal latent vector. To further ensure all the reconstructions obey the distribution of normal subsequences, we adversarially train the Autoencoder and discriminator, which will make the discriminator can not distinguish the reconstruction is from anomaly subsequences or normal subsequences and force Autoencoder to generate eligible reconstructions. By constraining both latent space and original space, we can better control the reconstruction process of anomalies and then obtain higher reconstruction error for anomalies. In the testing phase, the reconstruction error of test samples will be calculated as the anomaly score. The main contributions of this paper are as follows:

- Propose a new way of training the anomaly detection model, which can broaden the "vision" of the model. We use not only normal data but also the "imitated anomaly data" for model training.

- Provide a new thought for anomaly detection: ensure the reconstruction of normal and abnormal samples obey the distribution of normal samples and detect anomalies based on reconstruction error, which can improve the resolution between anomaly scores of anomaly subsequences and normal subsequences.

- To obtain richer features of subsequences, we use the fully convolutional network with different kernel size to extract neighbor information of datapoints in subsequences.

- Based on the above idea, we proposed a new deep-learning algorithm RAN for unsupervised time series anomaly detection by designing a specially designed and latent vector-constrained Autoencoder, and combining it with the discriminator to construct an adversarial network.

Extensive experiments on different types of time-series data sets from UCR Repository [14], BIDMC database [15, 16] and MIT-BIH datasets [17] show that (i) RAN can reconstruct normal subsequences well and ensure the reconstructions of anomaly subsequences obey the distribution of normal subsequences, which generates distinguishable reconstruction errors for anomaly detection. (ii) RAN can detect meaningful anomalies and get overall good performance in terms of AUC-ROC.

The remainder of this paper is organized as follows. Sect. 2 introduces some related work about unsupervised anomaly detection. Sect. 3 describes the details of the proposed unsupervised anomaly detection algorithm RAN. Experimental results and analysis are shown in Sect. 4. Finally, we conclude this paper in Sect. 5.

## 2 Related Work

### 2.1 Classical anomaly detection

Classical anomaly detection algorithms for time series data generally include two steps. First, they use different data structures to reduce the data dimension and represent data, then they calculate distance, density, or other statistic values based on the new representation to evaluate the anomaly degree. Distance-based algorithms [18, 19, 20] detects subsequences far away from other subsequences as anomalies. They represent time series by sequences or symbol sequences with reduced dimensions such as Piecewise Aggregate Approximation (PAA) [21] and Symbolic Aggregate Approximation (SAX) [22], but they need to calculate the pair-wise distance between subsequences, which usually lead to high time complexity. Density-based algorithms detect subsequences with low density as anomalies. They usually calculate local density based on neighborhoods such as Local Outlier Factor (LOF) [23] and Relative Density-based Outlier Score (RDOS) [24], but the performance is restricted with the number of neighborhoods. Isolation Forest(iForest) [25] is a particular algorithm that constructs isolation trees to represent data and detects samples with short average path lengths as anomalies. It works well in high dimensional datasets but may not fit for time series data, since it loses time-order information when selecting the data attributes randomly. There are also algorithms [26, 27] apply the hidden Markov model to detect samples with low probability as anomalies. They first calculate transition probabilities in different ways, then obtain the final probabilities by the iterative optimization process. However, the Markov model method also consumes lots of time for the iteration process.

### 2.2 Deep-learning anomaly detection

As the volume and dimension of data grow, more and more deep-learning algorithms are proposed for handling these complex data in anomaly detection. Most of these algorithms are based on the generative model and detect samples with high reconstruction error as anomalies. AnoGAN [8] is the first work that applies GAN for image anomaly detection. AnoGAN uses normal data to train the model and calculate errors from the trained generator and discriminator as anomaly scores. To decrease the test time and make the method more practical, [9] build ALAD upon bi-directional GANs and added an encoder network that maps data samples to latent variables. [28] first, apply Autoencoder(AE) for anomaly detection and also use reconstruction error to detect the anomaly. Considering that the reduced low-dimension space of AE is unable to preserve essential information, [10] proposed DAGMM by combining AE with a Gaussian Mixture Model(GMM) and adopting the joint training strategy. To mitigate the drawback that AE sometimes can also reconstruct anomalies well, [11] proposed MemAE which equipped the AE with a memory module to strengthen the reconstruction error of anomalies.

There are also some anomaly detection algorithms based on LSTM due to its special structure designed for handling time series data. LSTM-AD [29] is the first work applying LSTM for time-series anomaly detection. LSTM-AD trained the LSTM on normal data and use it as a predictor, then the prediction errors are used to model a multivariate Gaussian distribution, finally, the likelihood is used to evaluate anomaly degree. Considering that some time-series are unpredictable, [30] proposed EncDec-AD by constructing the Autoencoder with LSTM units. EncDec-AD tries to reconstruct normal time-series in the training phase, and use the reconstruction error to detect anomalies in the testing phase.

### 2.3 Novelty detection

Novelty detection also deserves paying attention, which is similar but different from anomaly detection. Both of them own the normal class data or target class data in training and try to separate other different data. Novelty detection is most in computer vision problems. it is identifying the new data outside the target classes, and these new data can be normal but just different classes. Anomaly detection pays more attention to abnormal behaviors, which can be generated in the same scenario such as the device operation. There are also some deep-learning algorithms for novelty detection based on generative models, which also inspire us more thinking. Due to the absence of novelty data, [13] proposed an end to end architecture ALOCC for novelty detection by combining the Autoencoder and discriminator. Considering that out-of-class samples can also be well represented in other novelty detection algorithms, [31] proposed a more complex model OCGAN to ensure the generated sample are from the known class.

## 3 Proposed method: RAN

This section will elaborate on the proposed method RAN which is based on the idea of Reconstruct Anomalies to Normal for unsupervised Time Series Anomaly Detection. Before introducing the algorithm, we will describe the time-series anomaly detection problem and present some symbols used later.

### 3.1 Problem description

Detecting anomaly subsequences is meaningful in real life. For example, detecting anomaly ECG subsequences can indicate health status and is necessary before the detailed heart disease diagnosis. Considering that anomalies always have different and uncertain lengths, it is more practical to first detect anomaly subsequences and then take a more detailed examination by experts under most circumstances.

For Time-series $X$ with $n$ subsequences $X = \{X_0, X_1, ..., X_{n-1}\}$ and the length of the subsequence is $m$, $X_i = \{x_{i\_0}, x_{i\_1}, ..., x_{i\_m-1}\}$. Assume we have the normal data set $X_{nor}$ used for model training and the test data set is $X_{test}$ which contains normal and abnormal subsequences. The object is to find anomaly subsequences in $X_{test}$.

### 3.2 Imitate anomaly subsequences

Most reconstruction-based anomaly detection algorithms only use normal data for model training, which ensures the good reconstruction of normal data but cannot control the reconstruction of anomalies. So, the reconstruction error of anomalies can sometimes be small and lead to the omission of anomalies. To broaden the "vision" of the model, we proposed a new way to train the anomaly detection model. We use not only normal data but also the "imitated anomaly data" for model training.

In most time-series anomaly detection scenarios, the anomaly part in an abnormal subsequence is usually a part rather than the whole. As is shown in Fig.1, the ECG record 308 is downloaded from the MIT-BIH ST Change Database(stdb). The anomaly subsequence is colored yellow, and the actual anomalies annotated by the expert are colored red. This anomaly subsequence also has normal data points and the anomaly is just the small unusual part.

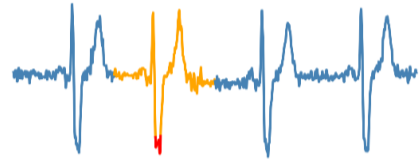

**Fig 1.** Anomalies in the anomaly subsequence

Based on the above observation, we corrupt the normal subsequence to imitate the anomaly subsequence. First, we randomly select some column indexes of the subsequence to get the index set $R$. Since anomaly data points do not conform to the distribution of normal data [32] and the normalized time series most have Gaussian distribution [33], we use Eq.(3-1) to calculate the corresponding unusual data values, in which $u_r$ and $\sigma_r$ are the mean value and variance of all $x_{i\_r}$ in the normal data set $X_{nor}$. Finally, we use these unusual data values as anomaly values to replace the normal data points and obtain the anomaly subsequence. The pseudo-code is shown in Algorithm 1.

$$ano\_x_r = u_r + 4 * \sigma_r \quad (r \in R) \quad (3\text{-}1)$$

As shown in Fig.2, the first row is the original subsequences and the second row is the "imitated" anomaly subsequences generated by the above steps. We corrupt the final two normal subsequences to imitate the anomaly subsequences and the anomaly data points are highlighted in red.

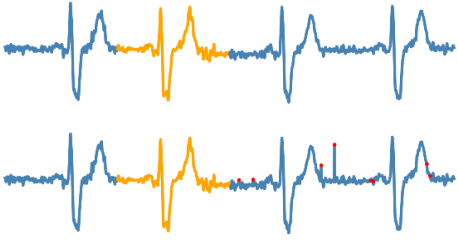

**Fig 2.** The original subsequences and imitated anomaly subsequences

---
**Algorithm 1**: Imitate anomaly subsequences
**Input:** the normal data set $X_{nor}$, corrupt level $c$.
**Output:** Imitate anomaly subsequences $X_{imi}$.
1  $n \leftarrow$ number of subsequences
2  $m \leftarrow$ length of the subsequence
3  $X_{imi} = X_{nor}$
4  **for** $X_i$ in $X_{imi}$ **do**
5     $R =$ randomly select $c*m$ indexes of $[1,2,\ldots m]$
6     **for** $r$ in $R$ **do**
7        $u_r = \frac{1}{n} * \sum_i^n x_{i\_r}(x_{i\_r} \in X_{nor})$
8        $\sigma_r = \frac{1}{n} * \sum_i^n (x_{i\_r} - u_r)(x_{i\_r} \in X_{nor})$
9        $x_{i\_r} = u_r + 4 * \sigma_r \ (x_{i\_r} \in X_{imi})$
10 **return** $X_{imi}$.

---

### 3.3 Reconstruct anomalies to normal

To best utilize the reconstruction error for anomaly detection, we aim to minimize the reconstruction error of the normal data and maximize the reconstruction error of anomalies as possible. In real life, the best way to judge real and fake is to compare the object with the standard real thing, and the detection anomalies are also the same. As shown in Fig.3, if we can use the normal data as the judging standard, and then calculate the error between this standard and the test object, we can get the higher error for anomalies and detect them more easily.

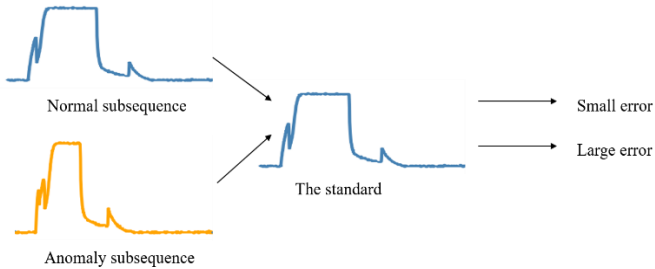

**Fig 3.** The original subsequences and imitated subsequences

Based on the above statement and inspired by the success of the generative model and adversarial learning, we migrate the architecture of ALOCC but improve it to complete our idea. Different from ALOCC to directly use outputs of discriminator as novelty probability, we use the reconstruction of AE as the standard and calculate the reconstruction error as the anomaly score. We also introduce the "imitating anomaly data" strategy and add constrain in the latent space. The key sight of our method is to ensure the reconstruction fits the distribution of normal data, which means reconstructing normal data well as possible and making the reconstruction of anomalies obey the distribution of normal data at the same time. In this way, we can get a higher reconstruction error for the anomaly subsequence than the normal subsequences and improve the detection accuracy.

The structure of our model is shown in Fig.4. $X_{nor}$ is the normal data we have, and $X_{imi}$ is the imitated anomaly data generated by steps in section 3.2. $Z$ is the latent vector of $X_{nor}$, and $Z_{imi}$ is the latent vector of $X_{imi}$ in the latent space. $X_{rec}$ is the reconstruction data generated by Decoder. To ensure the reconstruction of anomalies the same as normal, we minimize the latent vector error $Z_{\_error}$ as possible. To ensure the reconstruction fits the distribution of normal data, we add the discriminator Dx and use the AE as a generator to construct the adversary network, then the reconstruction will obey the same distribution with the normal data after adversarial training. By applying constraints both on the original space and the latent space of AE，we can force the model to learn the normal features better and ensure the reconstruction to be normal. Then, we can obtain distinguishable anomaly scores to better detect anomaly subsequences. More details about each component of the model are as follows:

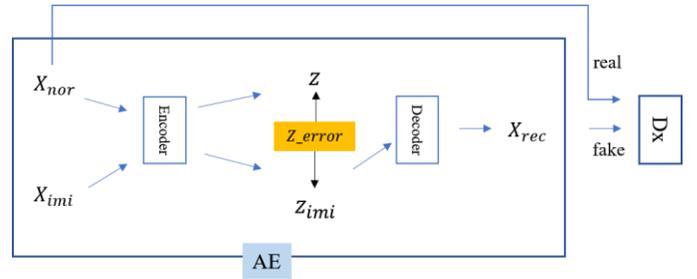

**Fig 4.** The structure of the proposed method

**Encoder**: Since the encoder is essential for generating a good latent-space representation, we specially design the structure of the encoder to extract better features for time series data. Datapoints in the subsequence are in time order and have relations with neighbors. So, data points combined with neighbors can contain more information. For example, a single low heartbeat point can be normal in the whole subsequence, while the continuous low heartbeats indicate anomaly circumstances. To extract richer information of subsequences, we use the 1D convolutional neural network as shown in Fig.5 to construct the encoder and set different kernel sizes in a different layer. To broaden the "vision" of the model, we also use the imitated anomaly subsequences $X_{imi}$ to feed the encoder. Then we can get the corresponding latent representation $Z_{imi}$.

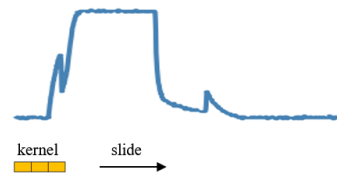

**Fig 5.** 1D-CNN to extract the combined information with neighbors.

**Decoder**: Decoder is another part of AE and we use the 1D deconvolutional neural network to construct it. We use the decoder to generate the reconstruction $X_{rec}$ from the latent representation $Z_{imi}$. The reasons why not use both $Z$ and $Z_{imi}$ are: 1) we minimize the error $Z_{error}$ between them during the training process; 2) we also force the reconstruction of $X_{imi}$ to have the same distribution with normal data in the adversarial training, and it will indirectly force $Z_{imi}$ have the same distribution with $Z$. So, the reconstruction of $Z$ and $Z_{imi}$ will be similar and we only need to use one of them.

**AE**: AE is the combination of encoder and decoder. AE also acts as the generator in the generative adversarial network. The aims of AE are: 1) for normal data, learning a good representation in the latent space and generating good reconstructions in the original data space; 2) for anomaly data, learning a representation as normal in the latent space and then generating reconstructions which obey the distribution of normal data in the original data space. We apply constraints both on latent space and the original data space to achieve these aims. The loss function for AE in the training process is as Eq. (3-2), and $\lambda = 10$ is given to the $Z_{error}$ term to obtain good reconstructions. The weight λ was chosen based on the quality of reconstruction. $Z_{error}$ is the loss between $Z$ and $Z_{imi}$ in the latent space, and $gen_{loss}$ is the loss in the original data space. Since we hope the reconstruction generated by AE obey the distribution of the normal data, we fool the discriminator Dx to make it can not distinguish the reconstruction and the normal data.

$$L_{AE} = \lambda * Z_{error} + gen_{loss}$$
$$Z_{error} = Z - Z_{imi} \qquad (3\text{-}2)$$
$$gen_{loss} = E_{X_{imi} \sim P_i}[\log(1 - Dx(AE(X\_imi)))]$$

**Dx**: Dx is the discriminator in the generative adversarial network and tries to distinguish the reconstruction and the normal data. Dx is trained to identity the reconstruction $X_{rec}$ as fake and the normal data $X_{nor}$ as real. The loss for Dx in the adversarial training procedure is as Eq. (3-3).

$$L_{Dx} = E_{X_{rec} \sim P_X}[\log(Dx(X_{rec}))] \qquad (3\text{-}3)$$

The pseudo-code of the proposed method **RAN** is shown in Algorithm 2.

---
**Algorithm 2**: Reconstruct anomalies to normal: **RAN**

Input: the normal data set $X_{nor}$, Imitated anomaly subsequences $X_{imi}$, the test data set $X_{test}$.
**Output:** reconstruction errors $rec\_errors$.
1  $n \leftarrow$ number of subsequences
2  $m \leftarrow$ length of the subsequence
3  Training phase:
3  **for** $epcho$ 1 to $N$ **do**
4      input $X_{nor}$ and $X_{imi}$ into AE
5      get the latent vector $Z$ and $Z_{imi}$, get output $X_{rec}$
5      Discriminator Dx update:
6      $L_{Dx} \leftarrow Dx(X_{rec}, 0) + Dx(X_{nor}, 1)$
7      Back-propagate $L_{Dx}$ and change Dx.
8      keep Dx fixed.
9      Generator AE update:
10     $Z_{error} = Z - Z_{imi}$
11     $gen_{loss} = Dx(X_{rec}, 1) + Dx(X_{nor}, 0)$
12     $L_{AE} = 10 * Z_{error} + gen_{loss}$
13     Back-propagate $L_{AE}$ and change AE.
14 Testing phase:
15 **for** $X_i$ in $X_{test}$:
16     keep model fixed.
17     $X_{i\_rec} = AE(X_i)$
18     $rec\_errors[i] = X_{i\_rec} - X_i$

---

**Anomaly detection**: After getting reconstruction errors of test subsequences, we use Eq. (3-4) to calculate anomaly scores based on them. A higher anomaly score indicates a higher possibility to be anomaly subsequence.

$$Ano\_Score_i = \frac{rec\_errors[i] - Min(rec\_errors)}{Max(rec\_errors) - Min(rec\_errors)} \qquad (3\text{-}4)$$

## 4 Experiment

In this section, we first introduce some data sets, then apply our algorithm and other typical anomaly detection algorithms to compare and analyze their performances. We also carried an ablation study to verify the effectiveness of each component in **RAN**.

### 4.1 Experiments Setup

**Data sets:** As shown in Table 1, four different types of time-series data sets are selected from the UCR Time Series Repository, MIT-BIH data sets, and BIDMC database to test the performance of these algorithms. These data sets are collected from different scenes. In ECG data, each subsequence traces the electrical activity recorded during one heartbeat. Anomalies in "ECG200" are heart attacks due to prolonged cardiac ischemia. Data in "BIDMC_chf07" are collected from a patient who has severe congestive heart failure. Anomalies in "MIT-BIH220" are atrial premature beats. Anomalies in "MIT-BIH221" are premature ventricular contraction beats. Anomalies in "MIT-BIH210" contains four types of abnormal beats('a', 'V', 'F', 'E'): atrial premature beats, premature ventricular contraction beats, the fusion of ventricular and normal beats, and ventricular escape beats. Sensor, motion, and image data sets are from the UCR Time Series Repository. Sensor data are collected from different sensors and divided into subsequences in the same time interval. Motion data is obtained according to the center of mass of the action. For image data, the contours of these images are extracted and mapped into a one-dimensional sequence from the center. There are several classes in some data sets, and considering that in real life the types of anomalies are often uncertain, we select one class as normal data and randomly select some samples from the other classes as anomaly data.

**Experimental Setup:** We select some typical anomaly detection algorithms for comparison. For classical anomaly detection algorithms, we select ESAX, SAX_TD, Interval, RDOS, PAPR, and iForest. For deep-learning anomaly detection algorithms, we select AnoGAN, DAGMM, ALAD, MemAE, LSTMAD, and LSTMED. All the above algorithms are used to calculate anomaly scores for all test samples. We implemented experiments on the computer server with 10 core CPU, 3.3GHz, 64 bits operation system. All codes are built in Python 3.7.

**Table 1.** The description of time-series data sets

| No. | data sets | seq_num | seq_length | ano_rate | types |
|---|---|---|---|---|---|
| 1 | ECG200 | 200 | 96 | 33.50% | ECG |
| 2 | BIDMC_chf07 | 5000 | 140 | 41.62% | ECG |
| 3 | MIT-BIH210 | 2649 | 207 | 8.57% | ECG |
| 4 | MIT-BIH220 | 2047 | 292 | 4.59% | ECG |
| 5 | MIT-BIH221 | 2426 | 191 | 16.32% | ECG |
| 6 | Lighting2 | 121 | 637 | 39.66% | Sensor |
| 7 | MoteStrain | 1272 | 84 | 46.14% | Sensor |
| 8 | SonyAIBORobotSurfaceII | 980 | 65 | 38.36% | Sensor |
| 9 | StarLightCurves | 427 | 1024 | 35.59% | Sensor |
| 10 | ToeSegmentation2 | 166 | 343 | 25.30% | Motion |
| 11 | GunPointAgeSpan | 339 | 150 | 32.74% | Motion |
| 12 | UWaveGestureLibraryX | 950 | 315 | 41.16% | Motion |
| 13 | DistalPhalanxOutlineCorrect | 876 | 80 | 38.47% | Image |
| 14 | HandOutlines | 1370 | 2709 | 36.13% | Image |
| 15 | DiatomSizeReduction | 142 | 345 | 30.99% | Image |

### 4.2 Experiments Analysis

**Performance Evaluation Methods:** Since most anomaly detection algorithms calculate anomaly scores to detect anomalies, we use the Area Under Receiver Operating Characteristic Curve (AUC-ROC) to have a comprehensive evaluation of these algorithms. In anomaly detection, higher AUC-ROC indicates a higher ability for the algorithm to distinguish anomaly subsequences and normal subsequences.

**Effectiveness to detect anomalies:** We select the "MIT-BIH210" from MIT-BIH Database to show that our algorithm can detect true anomalies. The "MIT-BIH210" data set contains five types of heartbeats, of which one type('N') is normal heartbeats and other types ('a', 'V', 'F', 'E') are anomaly heartbeats annotated by experts. A fragment of experiment results is shown in Fig. 6: The first row is test subsequences and anomaly subsequences are marked in orange color. The second row is the corresponding reconstructions and reconstructions of anomaly subsequences are marked in red color. The third row is the corresponding reconstruction errors of test subsequences. From these three subgraphs, we can see that our model can reconstruct the normal subsequences well and ensure the reconstruction of anomaly subsequences be similar to normal subsequences. Then, as shown in the third row, reconstruction errors of anomaly subsequences are higher than this of normal subsequences. Thus, we can detect anomalies more easily based on reconstruction errors.

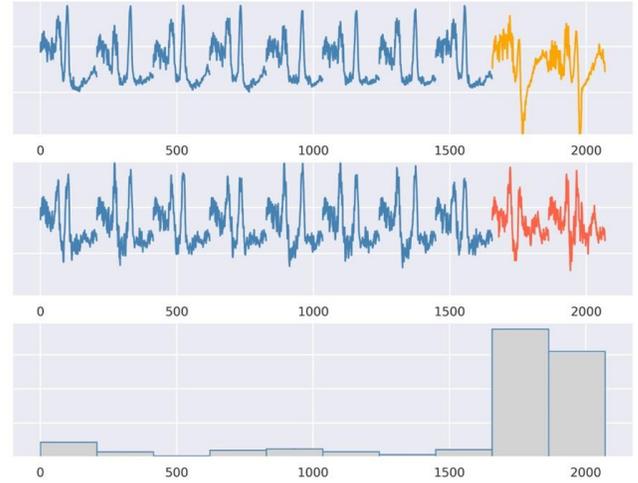

**Fig 6.** A fragment of the experiment results from MIT-BIH210.

**Improve the discrimination of anomaly scores:** Most anomaly detection algorithms output anomaly scores to determine anomalies, so it will be easier and more accurate to detect anomalies if we can improve the difference between anomaly scores of normal subsequences and anomaly subsequences. We show the histogram of anomaly scores from different algorithms in Fig.7. The anomaly scores of normal subsequences are colored in blue and the anomaly scores of anomaly subsequences are colored in red. The larger the overlap area, the harder to distinguish normal and anomaly subsequences. From Fig.7 we can find that anomaly scores generated by the proposed method **RAN** have the smallest overlap. Thus, the proposed method can improve the discrimination of anomaly scores compare to other methods.

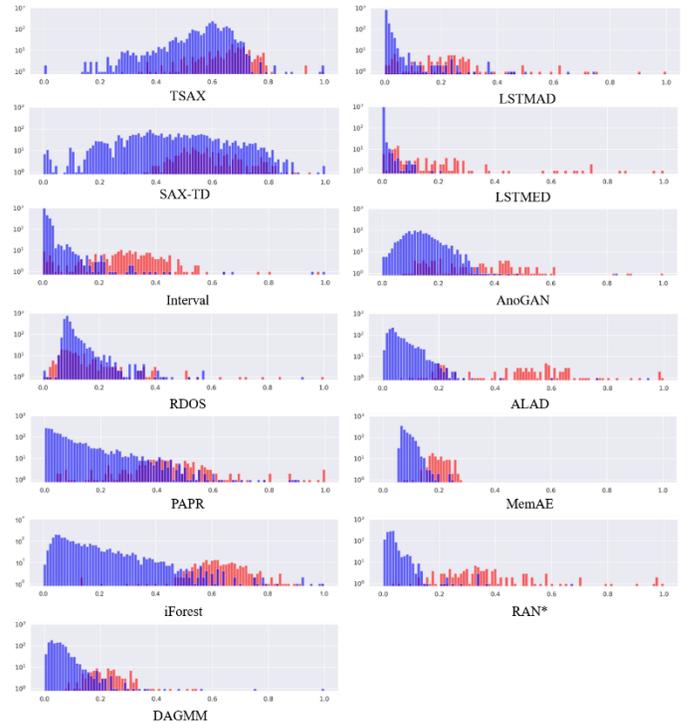

**Fig 7.** Anomaly scores from different algorithms.
\* is the proposed algorithm

**Accuracy:** Experimental results of the proposed algorithm and other algorithms are recorded in Table 2, and the best AUC-ROC are highlighted in bold font.

From Table 2, we can find that: 1) RAN outperform other algorithms in most data sets (9/15), which reflects the ability of RAN to detect anomalies for different types of time-series data; 2) MemAE obtains the second-best performance (4/15), which equips autoencoder with a memory module to mitigate the drawback of AE that it sometimes reconstruct the anomalies well. And it also reflects the importance of reconstructing anomalies to normal for reconstruction-based anomaly detection models. 3) Compare to non-deep-learning algorithms, deep-learning algorithms can get overall better performance due to their complex networks to extract more deep features, and they are more appropriate to process complex data.

**Table 2.** AUC-ROC of different algorithms.

The best AUC-ROC are highlighted in bold

| No | TSAX | SAX_TD | Interval | RDOS | PAPR | iForest | DAGMM | LSTMAD | LSTMED | AnoGAN | ALAD | MemAE | RAN* |
|---|---|---|---|---|---|---|---|---|---|---|---|---|---|
| 1 | 0.688 | 0.590 | 0.549 | 0.638 | 0.760 | 0.854 | 0.688 | 0.894 | 0.846 | 0.734 | 0.652 | 0.845 | **0.907** |
| 2 | 0.638 | 0.595 | 0.546 | 0.507 | 0.825 | 0.695 | 0.961 | 0.961 | 0.953 | 0.891 | 0.934 | 0.948 | **0.983** |
| 3 | 0.727 | 0.602 | 0.949 | 0.619 | 0.945 | 0.983 | 0.979 | 0.962 | 0.986 | 0.848 | 0.979 | 0.985 | **0.988** |
| 4 | 0.593 | 0.650 | 0.509 | 0.537 | 0.889 | 0.999 | 0.993 | 0.999 | 0.999 | 0.999 | 0.999 | **1.000** | **1.000** |
| 5 | 0.970 | 0.507 | 0.518 | 0.504 | 0.962 | 0.911 | 0.980 | 0.907 | 0.991 | 0.958 | 0.980 | 0.990 | **0.999** |
| 6 | 0.745 | 0.526 | 0.662 | 0.608 | 0.619 | **0.766** | 0.626 | 0.586 | 0.717 | 0.641 | 0.642 | 0.650 | 0.732 |
| 7 | 0.543 | 0.580 | 0.543 | 0.578 | 0.659 | 0.766 | 0.775 | 0.762 | 0.832 | 0.707 | 0.821 | **0.939** | 0.923 |
| 8 | 0.651 | 0.605 | 0.525 | 0.533 | 0.521 | 0.794 | 0.819 | **0.971** | 0.970 | 0.642 | 0.700 | 0.873 | 0.928 |
| 9 | 0.939 | 0.962 | 0.557 | 0.536 | 0.621 | 0.740 | 0.848 | 0.976 | 0.978 | **1.000** | **1.000** | **1.000** | **1.000** |
| 10 | 0.549 | 0.758 | 0.702 | 0.766 | 0.777 | 0.784 | **0.813** | 0.739 | 0.751 | 0.539 | 0.510 | 0.558 | 0.608 |
| 11 | 0.835 | 0.784 | 0.537 | 0.569 | 0.695 | 0.901 | 0.851 | 0.907 | 0.916 | 0.733 | 0.866 | 0.924 | **0.934** |
| 12 | 0.622 | 0.706 | 0.534 | 0.612 | 0.557 | 0.908 | 0.857 | 0.860 | 0.889 | 0.671 | 0.900 | 0.915 | **0.927** |
| 13 | 0.517 | 0.579 | 0.520 | 0.747 | 0.624 | 0.767 | **0.859** | 0.747 | 0.791 | 0.560 | 0.613 | 0.632 | 0.642 |
| 14 | 0.548 | 0.538 | 0.577 | 0.699 | 0.728 | 0.786 | 0.778 | 0.896 | **0.928** | 0.576 | 0.891 | 0.860 | 0.863 |
| 15 | 0.536 | 0.702 | 0.824 | 0.589 | 0.967 | 0.940 | 0.827 | **1.000** | 0.985 | **1.000** | **1.000** | **1.000** | **1.000** |

**Ablation Study:** We also carried the ablation study to verify the effectiveness of each component of the proposed model. We compare our model **RAN** with the following variants. In the testing phase, we set the same length of the latent vector and the same number of network layers for these models. In the testing phase, we calculate the reconstruction errors as anomaly scores.

**AE**: AE is the autoencoder constructed by fully connected networks and we only constrain the original data space by reducing the MSE of original data and the reconstruction in the training phase.

**AE-FCN**: AE-FCN is the autoencoder constructed by 1D-fully convolutional networks and we only constrain the original data space by reducing the MSE of original data and the reconstruction in the training phase.

**LAE-FCN**: LAE-FCN is latent-constrained AE-FCN which also constrains the latent space. In the training phase, we use the imitated anomaly data in LAE and reducing the MSE between $Z_{imi}$ and $Z_{nor}$ and the MSE between $X_{rec}$ and $X_{nor}$ as shown in Fig.4.

The AUC-ROC results of the ablation study are shown in Table 3. The proposed method RAN outperformance other variants in most datasets and gets overall good performance, which indicates that the proposed strategies are effective to improve the model performance.

**Table 3.** AUC-ROC of ablation study.

\* is the proposed algorithm and the best AUC-ROC are highlighted in bold

| No | Dataset | RAN* | LAE-FCN | AE-FCN | AE |
|---|---|---|---|---|---|
| 1 | ECG200 | **0.907** | 0.862 | 0.894 | 0.887 |
| 2 | BIDMC-chf07 | **0.983** | 0.950 | 0.945 | 0.947 |
| 3 | MIT-BIH210 | 0.988 | 0.986 | 0.983 | **0.989** |
| 4 | MIT-BIH220 | **1.000** | 0.999 | **1.000** | 0.999 |
| 5 | MIT-BIH221 | **0.999** | 0.994 | 0.990 | 0.993 |
| 6 | Lighting2 | 0.732 | 0.606 | **0.761** | 0.681 |
| 7 | MoteStrain | **0.923** | 0.903 | 0.889 | 0.907 |
| 8 | SonyAIBORobotSurfaceII | **0.928** | 0.896 | 0.902 | 0.906 |
| 9 | StarLightCurves | **1.000** | 0.996 | **1.000** | 0.990 |
| 10 | ToeSegmentation2 | **0.608** | 0.508 | 0.579 | 0.601 |
| 11 | GunPointAgeSpan | **0.934** | 0.530 | 0.511 | 0.530 |
| 12 | UwaveGestureLibraryX | 0.927 | 0.916 | 0.924 | **0.931** |
| 13 | DiatalphalanxOutlineCorrect | 0.642 | **0.730** | 0.764 | 0.727 |
| 14 | HandOutlines | 0.863 | 0.877 | 0.873 | **0.885** |
| 15 | DiatomSizeReduction | **1.000** | **1.000** | **1.000** | **1.000** |

To take a further look at the functionality of each component, we also observe the reconstructions from different variants. As shown in Fig 8, the first row is part of the original time series data from "MIT-BIH210" and anomaly subsequences are colored in orange. The following rows are the corresponding reconstructions of AE, AE-FCN, LAE-FCN, and the proposed model **RAN**. In the second row, we can find that AE can not reconstruct two crests well, especially the reconstruction marked by the circle is fuzzy. In the third row, AE-FCN can better reconstruct the first crest compare to AE, which might due to the convolutional network can extract rich shape and trend information by combining the data point with its neighbors. However, the reconstruction of anomaly subsequence marked by the circle in the third row is not similar to the distribution of others, which indicates that AE-FCN is not enough to constrain the reconstructions of anomalies sometimes. In the fourth row, LAE-FAN solved this problem by adding the constrain of latent space and can better limit the reconstruction of anomalies. As shown in the fourth row, all the above variants can not reconstruct the second crest(marked by the circle) well. In the last row, we can see that RAN can generate better reconstructions similar to the distribution of original normal subsequences. Thus, the adversary training strategy can force the model to learn more complex features and generate reconstructions which fit the distribution of normal subsequences.

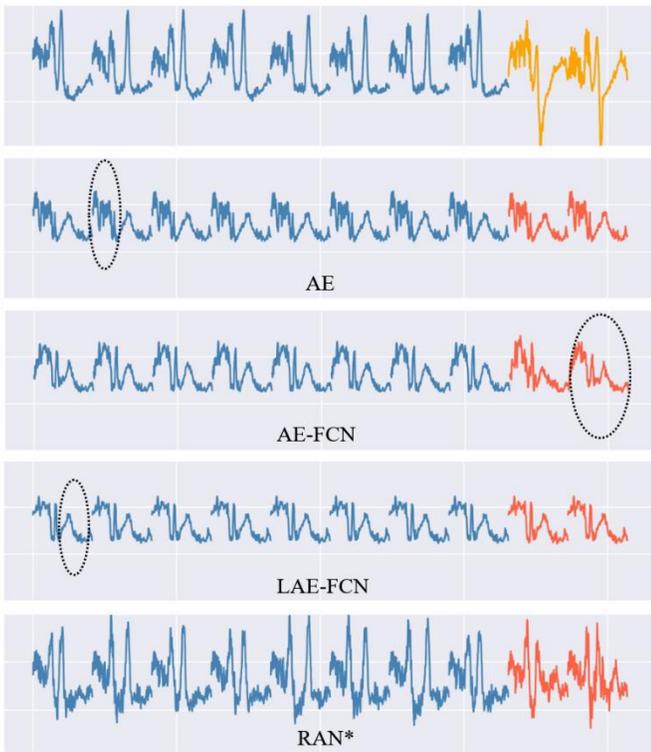

**Fig 8.** Original subsequences and the corresponding reconstructions from different models.

## 5 Conclusion

In this paper, a new method named reconstructing anomalies to normal (**RAN**) is proposed for detecting anomalies based on reconstruction errors. To fill the drawbacks that reconstruction-based algorithms only pay attention to reconstruct the normal data well, we proposed additionally control the reconstruction process of anomalies and ensure them obey the distribution of normal data. First, we imitate the anomaly data and feed them into the model to broaden the "vision" of the model. Then, we leverage the autoencoder as a generative model and construct it with 1D-fully convolutional networks to extract richer temporal information from data points and its neighbors. To ensure the reconstructions of both normal and anomaly data obey the distribution of normal data, we constrain both the latent space and original data space. In the latent space, we minimize the error between the latent vector of normal data and this of imitated anomaly data to guide the encoder to learn deep features and generate similar latent vectors. In the original space, we add the discriminator after the autoencoder and force the reconstructions to obey the distribution of normal data through adversarial learning. Finally, we can obtain more easily discriminable anomaly scores for test samples and get more accurate anomaly detection results.

Experimental results on diverse types of time series data sets also show that our algorithm **RAN** can detect meaningful anomalies and generate more easily discriminable anomaly scores than other algorithms. In terms of AUC-ROC, **RAN** also outperforms other algorithms on most datasets. The ablation study also shows that each component of RAN is meaningful and effective to improve the model performance.

## 6 Acknowledgment

This research is funded by Natural Science Foundation of Guangdong Province, China (Grant NO. 2020A1515010970) and Shenzhen Research Council (Grant NO. GJHZ20180928155209705).

## References

1. Chauhan, S., & Vig, L. (2015, October). Anomaly detection in ECG time signals via deep long short-term memory networks. In 2015 IEEE International Conference on Data Science and Advanced Analytics (DSAA) (pp. 1-7). IEEE.

2. Sivaraks, H., & Ratanamahatana, C. A. (2015). Robust and accurate anomaly detection in ECG artifacts using time series motif discovery. Computational and mathematical methods in medicine, 2015.

3. Ahmed, M., Mahmood, A. N., & Islam, M. R. (2016). A survey of anomaly detection techniques in financial domain. Future Generation Computer Systems, 55, 278-288.

4. Nicolau, M., & McDermott, J. (2018). Learning neural representations for network anomaly detection. IEEE transactions on cybernetics, 49(8), 3074-3087.

5. Pang, G., Cao, L., Chen, L., & Liu, H. (2018, July). Learning representations of ultrahigh-dimensional data for random distance-based outlier detection. In Proceedings of the 24th ACM SIGKDD International Conference on Knowledge Discovery & Data Mining (pp. 2041-2050).

6. Zimek, A., Schubert, E., & Kriegel, H. P. (2012). A survey on unsupervised outlier detection in high-dimensional numerical data. Statistical Analysis and Data Mining: The ASA Data Science Journal, 5(5), 363-387.

7. Chalapathy, R., & Chawla, S. (2019). Deep learning for anomaly detection: A survey. arXiv preprint arXiv:1901.03407.

8. Schlegl, T., Seeböck, P., Waldstein, S. M., Schmidt-Erfurth, U., & Langs, G. (2017, June). Unsupervised anomaly detection with generative adversarial networks to guide marker discovery. In International conference on information processing in medical imaging (pp. 146-157). Springer, Cham.

9. Zenati, H., Romain, M., Foo, C. S., Lecouat, B., & Chandrasekhar, V. (2018, November). Adversarially learned anomaly detection. In 2018 IEEE International Conference on Data Mining (ICDM) (pp. 727-736). IEEE.

10. Zong, B., Song, Q., Min, M. R., Cheng, W., Lumezanu, C., Cho, D., & Chen, H. (2018, February). Deep autoencoding gaussian mixture model for unsupervised anomaly detection. In International Conference on Learning Representations.

11. Li, D., Chen, D., Jin, B., Shi, L., Goh, J., & Ng, S. K. (2019, September). MAD-GAN: Multivariate anomaly detection for time series data with generative adversarial networks. In International Conference on Artificial Neural Networks (pp. 703-716). Springer, Cham.

12. Gong, D., Liu, L., Le, V., Saha, B., Mansour, M. R., Venkatesh, S., & Hengel, A. V. D. (2019). Memorizing normality to detect anomaly: Memory-augmented deep autoencoder for unsupervised anomaly detection. In Proceedings of the IEEE International Conference on Computer Vision (pp. 1705-1714).

13. Sabokrou, M., Khalooei, M., Fathy, M., & Adeli, E. (2018). Adversarially learned one-class classifier for novelty detection. In Proceedings of the IEEE Conference on Computer Vision and Pattern Recognition (pp. 3379-3388).

14. Chen, Y., Keogh, E., Hu, B., Begum, N., Bagnall, A., Mueen, A., & Batista, G. (July 2015). The UCR time series classification archive, http://www.cs.ucr.edu/eamonn/time_series_data/

15. Baim, D.S., Colucci, W.S., Monrad, E.S., Smith, H.S., Wright, R.F., Lanoue, A.S., Gauthier, D.F., Ransil, B.J., Grossman, W., & Braunwald, E. (1986). Survival of patients with severe congestive heart failure treated with oral milrinone. Journal of the American College of Cardiology 7(3), 661–670. https://doi.org/10.1016/S0735-1097(86)80478-8

16. Goldberger, A.L., Amaral, L.A.N., Glass, L., Hausdorff, J.M., Ivanov, P.C., Mark, R.G., Mietus, J.E., Moody, G.B., Peng, C., & Stanley, H.E. (2000). Physiobank, physiotoolkit, and physionet components of a new research resource for complex physiologic signals. Circulation 101(23), 215–220.

17. Moody, G.B., & Mark, R.G. (2001). The impact of the MIT-BIH arrhythmia database. IEEE Engineering in Medicine and Biology Magazine 20(3), 45–50. https://doi.org/10.1109/51.932724

18. Zhang, C., Chen, Y., Yin, A., & Wang, X. (2019). Anomaly detection in ECG based on trend symbolic aggregate approximation. Mathematical Biosciences and Engineering, 16(4), 2154-2167.

19. Sun, Y., Li, J., Liu, J., Sun, B., & Chow, C. (2014). An improvement of symbolic aggregate approximation distance measure for time series. Neurocomputing, 138, 189-198.

20. Ren, H., Liu, M., Liao, X., Liang, L., Ye, Z., & Li, Z. (2018). Anomaly detection in time series based on interval sets. IEEJ Transactions on Electrical and Electronic Engineering 13(5), 757–762. https://doi.org/10.1002/tee.22626

21. Keogh, E., Chakrabarti, K., Pazzani, M., & Mehrotra, S. (2001). Dimensionality reduction for fast similarity search in large time series databases. Knowledge and information Systems, 3(3), 263-286.

22. Lin, J., Keogh, E., Lonardi, S., & Chiu, B. (2003, June). A symbolic representation of time series, with implications for streaming algorithms. In Proceedings of the 8th ACM SIGMOD workshop on Research issues in data mining and knowledge discovery (pp. 2-11).

23. Breunig, M. M., Kriegel, H. P., Ng, R. T., & Sander, J. (2000, May). LOF: identifying density-based local outliers. In Proceedings of the 2000 ACM SIGMOD international conference on Management of data (pp. 93-104).

24. Tang, B., & He, H. (2017). A local density-based approach for outlier detection. Neurocomputing, 241, 171-180.

25. Liu, F. T., Ting, K. M., & Zhou, Z. H. (2008, December). Isolation forest. In 2008 Eighth IEEE International Conference on Data Mining (pp. 413-422). IEEE.

26. Moonesignhe, H. D. K., & Tan, P. N. (2006, November). Outlier detection using random walks. In 2006 18th IEEE International Conference on Tools with Artificial Intelligence (ICTAI'06) (pp. 532-539). IEEE.

27. Ren, H., Liu, M., Li, Z., & Pedrycz, W. (2017). A piecewise aggregate pattern representation approach for anomaly detection in time series. knowledge-based Systems, 135, 29-39.

28. Hawkins, S., He, H., Williams, G., & Baxter, R. (2002, September). Outlier detection using replicator neural networks. In International Conference on Data Warehousing and Knowledge Discovery (pp. 170-180). Springer, Berlin, Heidelberg.

29. Malhotra, P., Vig, L., Shroff, G., & Agarwal, P. (2015, April). Long short term memory networks for anomaly detection in time series. In Proceedings (Vol. 89, pp. 89-94). Presses universitaires de Louvain.

30. Malhotra, P., Ramakrishnan, A., Anand, G., Vig, L., Agarwal, P., & Shroff, G. (2016). LSTM-based encoder-decoder for multi-sensor anomaly detection. arXiv preprint arXiv:1607.00148.

31. Perera, P., Nallapati, R., & Xiang, B. (2019). Ocgan: One-class novelty detection using gans with constrained latent representations. In Proceedings of the IEEE Conference on Computer Vision and Pattern Recognition (pp. 2898-2906).

32. Chandola, V. , Banerjee, A. , & Kumar, V. . (2009). Anomaly detection: a survey. Acm Computing Surveys, 41(3), 1-3.

33. Lin, J. , Keogh, E. , Wei, L. , & Lonardi, S. . (2007). Experiencing sax: a novel symbolic representation of time series. Data Mining & Knowledge Discovery, 15(2), 107-144.